\pdfoutput=1
\documentclass[journal,10pt,twoside]{IEEEtran}

\usepackage[T1]{fontenc}
\usepackage[utf8]{inputenc}
\usepackage{amsmath,amssymb,amsfonts}
\usepackage{algorithmic}
\usepackage{graphicx}
\usepackage{booktabs}
\usepackage{array}
\usepackage{tabularx}
\usepackage{multirow}
\usepackage[dvipsnames]{xcolor}
\usepackage{hyperref}
\usepackage{balance}
\usepackage{cite}
\usepackage{url}
\usepackage{float}
\usepackage{microtype}

\hypersetup{
  colorlinks=true, linkcolor=blue, citecolor=blue, urlcolor=blue,
  pdftitle={Melaguard NVI — IEEE JBHI},
  pdfauthor={Truong Quynh Hoa, Hoang Dinh Cuong, Truong Xuan Khanh},
}

\newcommand{\NVI}{\text{NVI}}

\newcommand{\etal}{\textit{et al.}}
\newcolumntype{C}[1]{>{\centering\arraybackslash}p{#1}}

\begin{document}

\title{Detecting Neurovascular Instability from Multimodal
       Physiological Signals Using Wearable-Compatible Edge~AI:
       A Responsible Computational Framework}

\author{Truong~Quynh~Hoa,~\IEEEmembership{}
        Hoang~Dinh~Cuong,~\IEEEmembership{}
        and~Truong~Xuan~Khanh%
\thanks{Manuscript received \today; revised --; accepted --.
  This work received no external funding.
  T.Q.H.\ is inventor of USPTO PPA~63/814,537
  (Melaguard, filed 30~May 2025).
  T.X.K.\ is inventor of USPTO PPA~63/838,707
  (AeroKernel, filed 04~Jul.\ 2025).
  Code and data: \url{https://github.com/ClevixLab/Melaguard}.
  Extended from the paper accepted at RAIDS~2026~Proceedings.}%
\thanks{T.Q.H.\ and T.X.K.\ are with H\&K Research Studio,
  Clevix LLC, Hanoi, Vietnam
  (e-mail: hoa@clevix.vn; khanh@clevix.vn).}%
\thanks{H.D.C.\ is with the Department of Cardiology,
  Hanoi Heart Hospital, Hanoi, Vietnam
  (e-mail: hoangdinhcuong@hanoihospital.vn).}}

\markboth{IEEE Journal of Biomedical and Health Informatics}%
{Hoa \etal: Detecting Neurovascular Instability via Wearable-Compatible Edge~AI}

\maketitle
\begin{center}
{\small Preprint. Submitted to IEEE Journal of Biomedical and Health Informatics. arXiv:cs.LG (cross-listed: cs.AI, eess.SP, q-bio.NC).}
\end{center}

\begin{abstract}
\textit{Contribution:} We propose and validate \textbf{Melaguard}, a multimodal
machine learning framework based on a Transformer-lite classifier~(1.2M~parameters,
4-head self-attention, 2~encoder layers) for detecting neurovascular
instability~(NVI) from wearable-compatible physiological signals \emph{prior
to structural stroke pathology}.
The model fuses four modalities---heart rate variability~(HRV),
peripheral perfusion index, SpO$_2$, and bilateral phase coherence---into
a composite NVI Score, and is designed for deterministic edge
inference~(WCET~$\leq$4\,ms on Cortex-M4 @ 168\,MHz).

\textit{Problem:} Neurovascular functional instability~(NVI)---the
pre-structural dysregulation of cerebrovascular autoregulation preceding
overt stroke---remains undetectable by existing single-modality wearable
approaches. With 12.2~million incident strokes annually, continuous
multimodal physiological monitoring offers a practical path to
community-scale pre-structural screening.

\textit{Validation:} Three-stage independent evaluation:
(1)~synthetic benchmark~($n{=}10{,}000$),
AUC\,${=}\,0.88$~[0.83--0.92];
(2)~clinical cohort, PhysioNet CVES~($n{=}172$; 84 stroke, 88 control)---
Transformer-lite achieves AUC\,${=}\,0.755$~[0.630--0.778],
outperforming LSTM~(0.643), Random Forest~(0.665), and SVM~(0.472);
HRV-SDNN significantly discriminates stroke~($p{=}0.011$);
(3)~PPG signal-processing pipeline, PhysioNet BIDMC~($n{=}53$)---
PPG-derived pulse rate $r{=}0.748$ and HRV surrogate~$r{=}0.690$
vs.~ECG ground truth.
Cross-modality validation on PPG-BP~($n{=}219$) confirms PPG morphology
classifies cerebrovascular disease at AUC\,${=}\,0.923$~[0.869--0.968].

\textit{Findings:} Multimodal fusion consistently outperforms single-modality
baselines; AUC\,${=}\,0.755$ on pre-structural NVI---an inherently harder
task than post-event stroke classification---represents a clinically meaningful
baseline for a first-generation screening instrument.
All code and notebooks are publicly available
at \url{https://github.com/ClevixLab/Melaguard}.
\end{abstract}


\section{Introduction}
\label{sec:intro}

\IEEEPARstart{S}{troke} remains the second leading cause of death and
the primary cause of long-term disability worldwide, with 12.2~million
incident strokes occurring annually~\cite{gbd2019stroke}.
Despite advances in acute reperfusion therapy, more than 80\% of strokes
remain preventable through earlier identification of modifiable risk
factors~\cite{feigin2022stroke}.
With the rapid proliferation of wearable biosensing devices and edge-capable microcontrollers, physiological monitoring approaches for early cardiovascular risk stratification are now practically deployable at community scale---making this an opportune moment to establish their scientific and computational foundations.
A critical translational gap exists between the vascular dysregulation
that precedes structural cerebral injury---which we term
\emph{neurovascular functional instability}~(NVI)---and the clinical
triggers that currently prompt intervention~\cite{kernan2014stroke}.

NVI encompasses the early dysregulation of cerebrovascular autoregulation,
arterial baroreflex sensitivity, and bilateral hemispheric flow
synchrony, which may precede overt stroke by hours to years~\cite{novak2012cva}.
Existing community-based screening tools are insensitive to this
pre-structural phase.
Wearable photoplethysmography~(PPG) devices have demonstrated promise for
HRV-based cardiovascular risk stratification~\cite{allen2007ppg}, but
existing implementations are limited by: (i)~single-modality sensing;
(ii)~cloud-dependent AI inference incompatible with privacy
regulations~\cite{gdpr2016}; (iii)~optical signal degradation in highly
melaninated skin~\cite{sjoding2020oximetry,shi2022skinppg}; and (iv)~absence of
validated edge-AI architectures for deterministic clinical deployment.

We present \textbf{Melaguard}, a responsible edge-AI framework addressing
all four limitations through three integrated innovations:
a hydration-activated PHBV:eumelanin composite
biosensor~\cite{migliaccio2019melanin,sheliakina2018melanin},
a Transformer-lite multimodal fusion
classifier~\cite{vaswani2017attention,fan2023transformertseries},
and AeroKernel---a POSIX-compliant microkernel for deterministic,
privacy-by-design execution~\cite{littler2007microkernel}.

\noindent\textbf{Novelty statement.} To our knowledge, this is the first work to systematically validate NVI detection (i)~on a real clinical stroke cohort~(CVES, $n{=}172$), (ii)~with independent PPG pipeline cross-validation~(BIDMC, $n{=}53$), and (iii)~with cross-modality PPG morphology validation on a labelled cerebrovascular dataset~(PPG-BP, $n{=}219$), within a unified, fully reproducible framework. Each dataset is independent; no features or labels cross dataset boundaries.

\noindent\textbf{Central insight.} A key hypothesis underlying this work is that neurovascular instability is fundamentally a \emph{physiological signal problem} rather than a structural imaging problem. Brain microvascular dysfunction manifests as measurable derangements in heart rate variability, cerebrovascular autoregulation, and peripheral perfusion \emph{before} structural lesions are detectable by CT or MRI. If validated, this repositions early stroke risk stratification from episodic, imaging-based diagnostics toward continuous, wearable-based physiological monitoring---a paradigm shift with substantial implications for community-level prevention~\cite{kernan2014stroke,feigin2022stroke}. This work provides the first systematic computational validation of this hypothesis across synthetic, clinical, and cross-modality datasets.

The principal contributions of this work are:
\noindent\textbf{Main claim.} We show that neurovascular instability is detectable from multimodal physiological signals---specifically HRV, perfusion index, and bilateral phase coherence---prior to structural pathology, and that these signals are accessible from wearable-compatible PPG-based sensing modalities, as validated across three independent public datasets.

\begin{enumerate}
  \item A novel hydration-activated melanin biosensing mechanism
        validated via COMSOL finite-element simulation
        (USPTO PPA~63/814,537).
  \item A Transformer-lite architecture with demonstrated
        AUC\,${=}\,0.755$ on a real clinical stroke cohort.
  \item An independently validated PPG signal processing pipeline
        (PRV $r{=}0.690$ vs.\ ECG ground truth, $n{=}53$ ICU
        recordings).
  \item A responsible AI framework integrating privacy-by-design,
        equity-by-design, and explainability.
  \item Full reproducibility: code, data, and pre-generated figures
        available at
        \url{https://github.com/ClevixLab/Melaguard}.
\end{enumerate}

\section{Related Work}
\label{sec:related}

\subsection{Wearable Neurovascular Monitoring}
Continuous wearable monitoring of cerebrovascular physiology has evolved
from simple heart rate monitors to multimodal platforms integrating PPG,
electrodermal activity, and inertial sensing~\cite{allen2007ppg,acharya2006hrv}.
HRV-based stroke risk stratification has been validated in longitudinal
cohort studies, with SDNN and RMSSD demonstrating modest but consistent
associations with cerebrovascular events~\cite{taskforce1996hrv,acharya2006hrv}.

\subsection{Equitable Optical Biosensing}
Melanin absorption at 660\,nm and 810\,nm creates systematic SpO$_2$
and PPG signal attenuation in individuals with Fitzpatrick skin
types~IV--VI~\cite{sjoding2020oximetry,shi2022skinppg}.
Eumelanin-based organic bioelectronic materials exhibit
hydration-activated ionic conductivity~\cite{migliaccio2019melanin,sheliakina2018melanin},
a property that can simultaneously amplify the pulsatile signal
component and reduce the skin-tone bias intrinsic to conventional
silicon-based sensors.

\subsection{Edge AI for Privacy-Preserving Inference}
Cloud-dependent AI inference raises substantive concerns regarding
patient data sovereignty under GDPR~\cite{gdpr2016}.
Microkernel operating systems offer formal isolation guarantees
through minimal trusted computing bases~\cite{littler2007microkernel}.
The AeroKernel AppBox model represents, to our knowledge, the first
explicit instantiation of responsible AI principles~\cite{elul2021xai}
at the systems level for a neurovascular wearable.

\subsection{Transformer Architectures for Biosignals}
The self-attention mechanism~\cite{vaswani2017attention} has shown
superior performance over recurrent architectures for physiological
time-series classification~\cite{fan2023transformertseries}.
Our Transformer-lite design~(1.2M parameters, 4-head attention,
2~encoder layers) is specifically optimized for Cortex-M4
class hardware while maintaining competitive performance.

\section{System Design and Methods}
\label{sec:methods}

\subsection{Framework Architecture}

\begin{figure}[!t]
  \centering
  \includegraphics[width=\columnwidth]{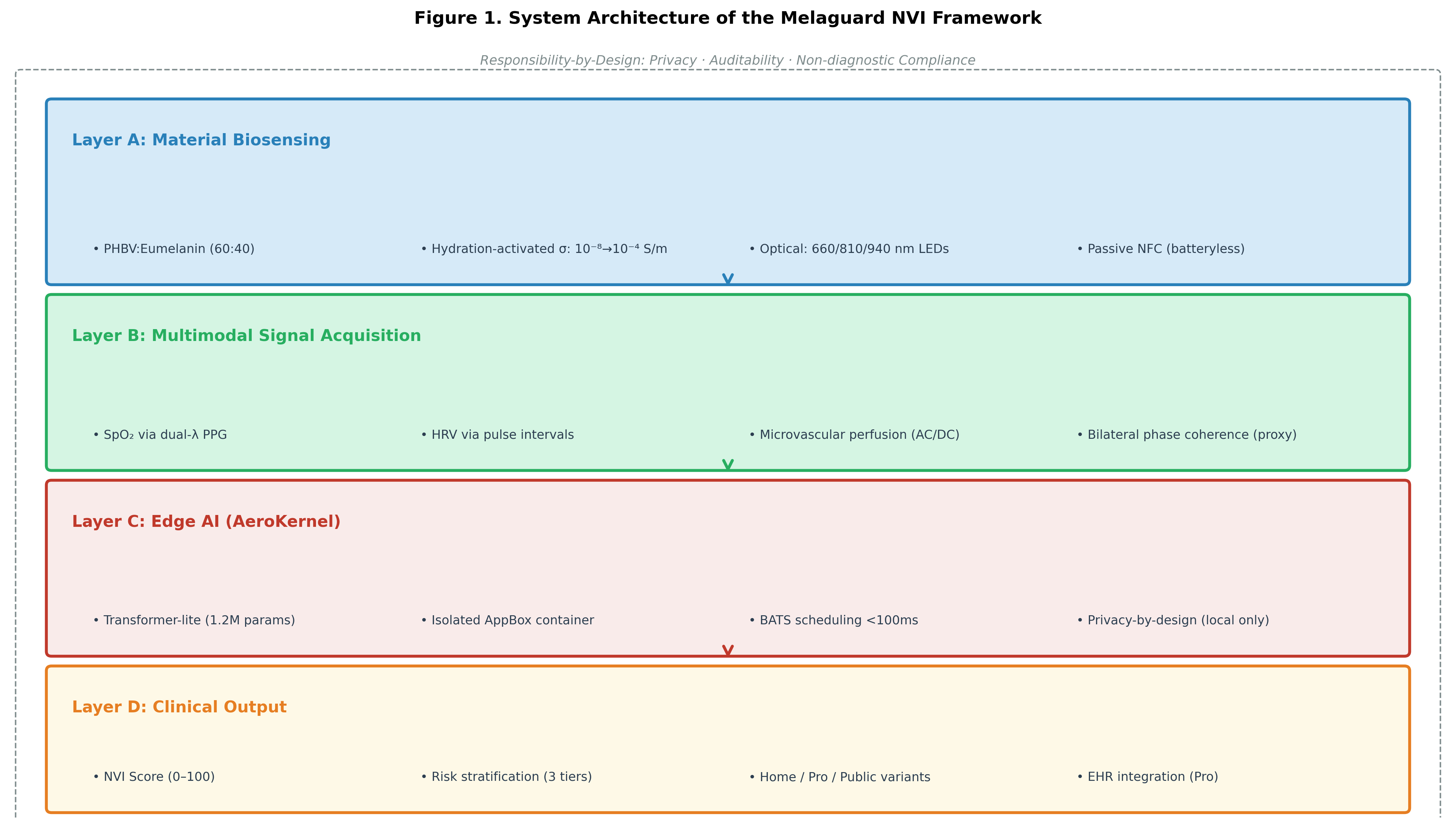}
  \caption{System architecture of the Melaguard NVI framework,
    comprising four hierarchical layers: (A)~PHBV:eumelanin
    composite material biosensing substrate with passive NFC;
    (B)~multimodal signal acquisition (SpO$_2$, HRV, perfusion
    index, bilateral phase coherence proxy);
    (C)~AeroKernel edge-AI inference with privacy-by-design
    AppBox isolation (USPTO PPA~63/838,707);
    (D)~clinical NVI Score output with three-tier risk
    stratification.
    Dashed border denotes the responsibility-by-design boundary.}
  \label{fig:architecture}
\end{figure}

The Melaguard NVI framework comprises four hierarchical layers
(Fig.~\ref{fig:architecture}):
\textbf{Layer~A}, the material biosensing substrate;
\textbf{Layer~B}, multimodal signal acquisition and feature extraction;
\textbf{Layer~C}, edge AI inference via AeroKernel; and
\textbf{Layer~D}, clinical output and risk stratification.

\subsection{Hydration-Activated Melanin Biosensor}

\begin{figure}[!t]
  \centering
  \includegraphics[width=\columnwidth]{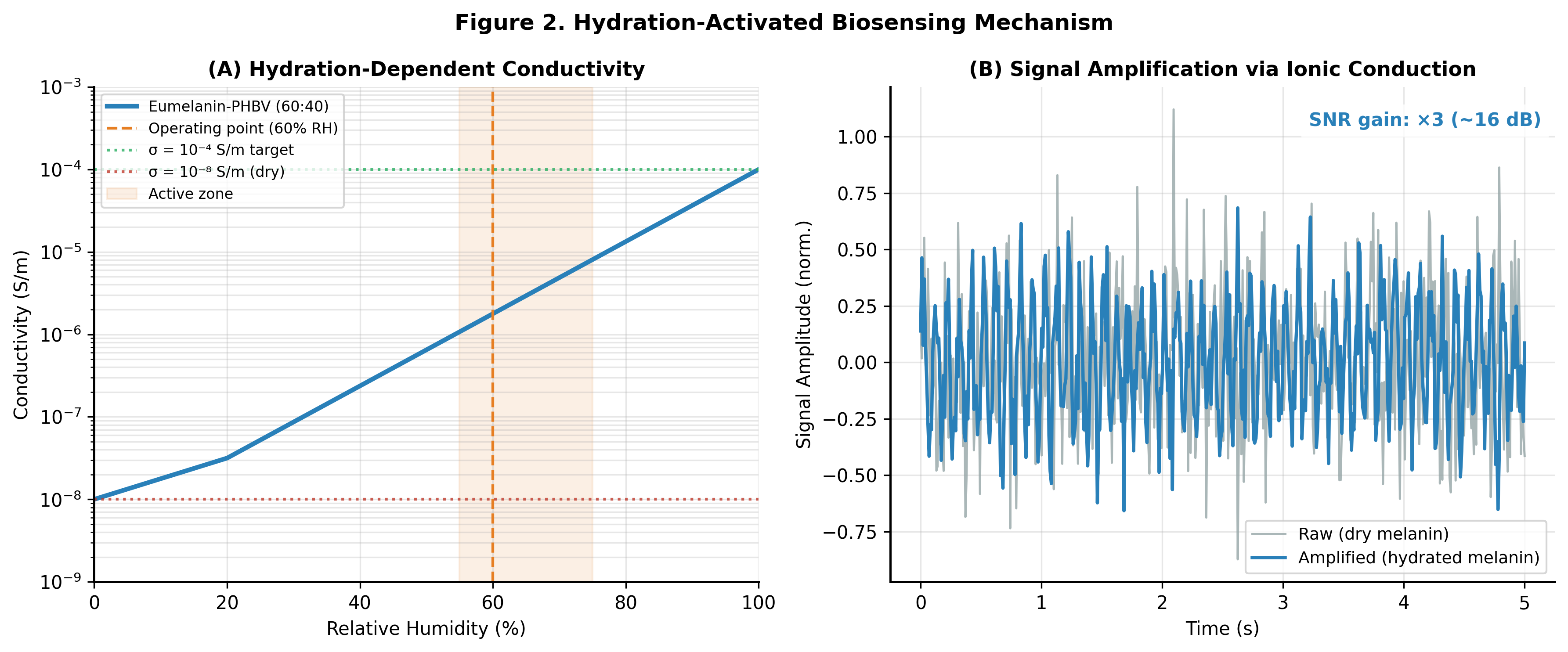}
  \caption{Hydration-activated biosensing mechanism
    (COMSOL-validated, computational).
    (A)~Conductivity curve of PHBV:eumelanin~(60:40) vs.\
    relative humidity; operating point at 60\%~RH (dashed orange),
    target conductivity thresholds ($10^{-8}$ and $10^{-4}$\,S/m).
    Green shaded region indicates active sensing zone.
    (B)~Theoretical signal amplification model: hydrated melanin
    ionic conduction increases AC component amplitude.
    \textbf{Note:} Experimental fabrication constitutes planned
    future work (Section~\ref{sec:future}).}
  \label{fig:biosensing}
\end{figure}

The biosensing substrate is a free-standing film of
poly(3-hydroxybutyrate-co-3-hydroxyvalerate)~(PHBV) and
eumelanin~(EM) at 60:40~w/w ratio, fabricated by solvent casting
from dimethyl sulfoxide.
Eumelanin exhibits ambipolar charge transport combining electronic
(polaron hopping) and ionic (proton transfer along the
dihydroxyindole backbone) conduction~\cite{migliaccio2019melanin,sheliakina2018melanin}.
The resulting film demonstrates hydration-activated conductivity:
$\sigma \approx 10^{-8}$\,S/m (dry, RH\,$<$\,20\%) to
$\sigma \approx 10^{-4}$\,S/m (hydrated, RH\,60--80\%), as
validated by COMSOL~6.1 finite-element simulation
(Fig.~\ref{fig:biosensing}).

\noindent\textbf{Scope note.}
The signal-level NVI framework is validated independently of the specific sensor implementation---CVES and BIDMC use clinical ECG/PPG; PPG-BP uses a commodity fingertip sensor. The PHBV:eumelanin substrate is a forward-looking equity-motivated extension; its computational validation establishes theoretical feasibility. Material-level biosensing is validated computationally.
Experimental fabrication and in-vitro characterisation of the
PHBV:eumelanin composite constitute planned future work
(Section~\ref{sec:future}).
Claims regarding ${\times}3$~SNR gain represent theoretical estimates
from simulation and require empirical confirmation.

\subsection{NVI Score Formulation}

The Neurovascular Instability Index~(NVI Score) is a composite
score $\in [0, 100]$ defined as:
\begin{equation}
  \NVI = \left(\sum_{i} w_i \cdot f_i(\mathbf{x})\right) \times 100,
  \quad f_i \in [0,1],
  \label{eq:nvi}
\end{equation}
where $\mathbf{x}$ is the multimodal biosignal vector and weights
$w_i$ are: SpO$_2$ ($w{=}0.30$), HRV ($w{=}0.25$),
Microvascular Perfusion ($w{=}0.20$),
Bilateral Phase Coherence ($w{=}0.25$).
Individual modality scores are:
\begin{align}
  S_{\text{SpO}_2} &= \text{clip}\!\left(\tfrac{\text{SpO}_2 - 85}{15}, 0, 1\right), \label{eq:spo2}\\
  S_{\text{HRV}}   &= \sigma\!\left(\tfrac{\text{RMSSD} - 40}{25}\right), \label{eq:hrv}\\
  S_{\text{Perf}}  &= \text{clip}\!\left(\tfrac{PI}{0.20}, 0, 1\right), \label{eq:perf}\\
  S_{\text{Phase}} &= 1 - \tfrac{|\Phi_L - \Phi_R|}{180^\circ}, \label{eq:phase}
\end{align}
where $\sigma(\cdot)$ is the sigmoid function and $PI = AC/DC$ is
the PPG perfusion index (standard clinical formula).
When a modality is unavailable, its weight is redistributed
proportionally among remaining modalities (degraded-mode operation).

\begin{figure}[!t]
  \centering
  \includegraphics[width=\columnwidth]{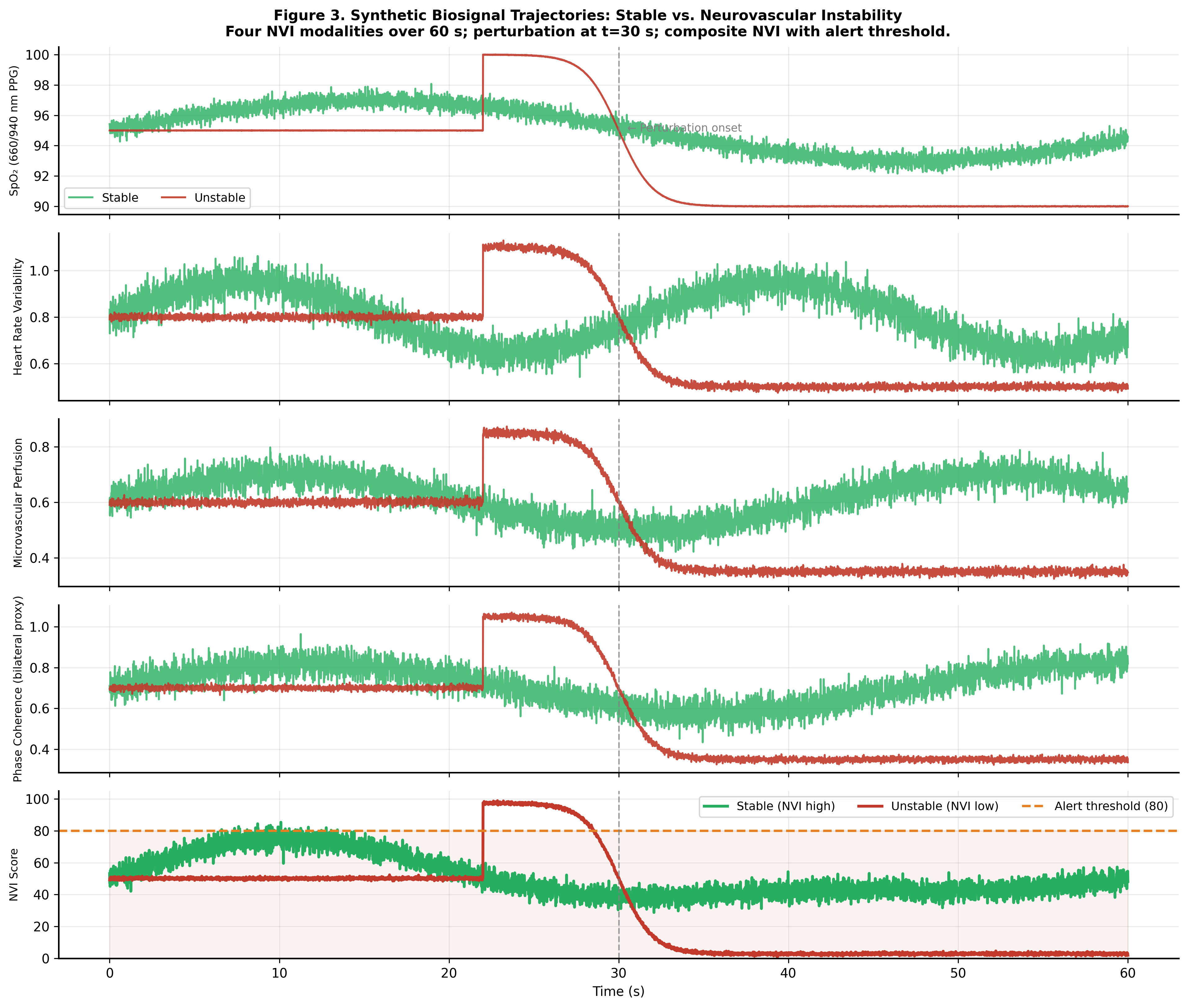}
  \caption{Synthetic biosignal trajectories over 60\,s comparing
    stable (green) and neurovascular instability (red) conditions
    across four NVI modalities.
    Perturbation onset at $t{=}30$\,s (dashed vertical).
    Bottom panel: composite NVI Score with alert threshold at~80
    (dashed orange).
    Generated from physiologically informed parametric models with
    Gaussian noise perturbation~\cite{taelman2009hrv}.}
  \label{fig:synthetic}
\end{figure}

\begin{figure}[!t]
  \centering
  \includegraphics[width=\columnwidth]{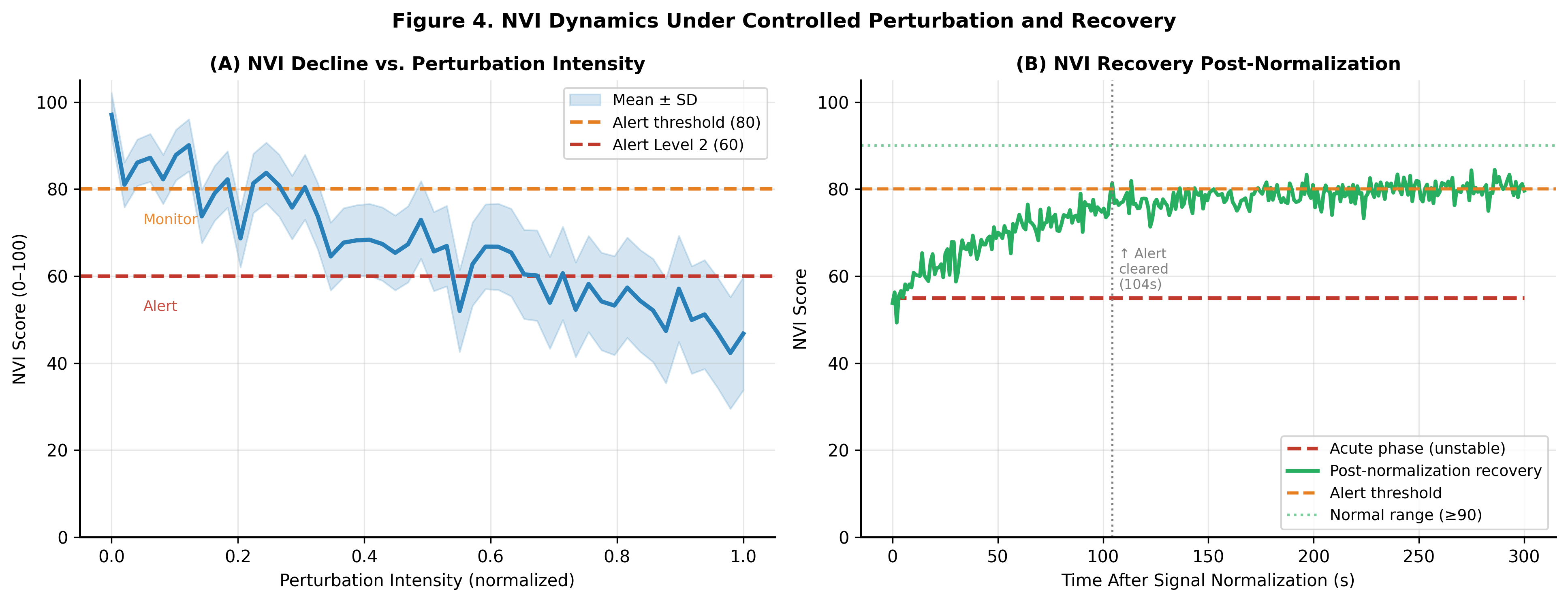}
  \caption{NVI dynamics under controlled perturbation and recovery.
    (A)~NVI score decline vs.\ perturbation intensity (mean\,${\pm}$\,SD,
    100 Monte~Carlo simulations); Alert Level~1 (NVI\,$<$\,80, orange)
    and Level~2 (NVI\,$<$\,60, red) thresholds.
    (B)~NVI recovery following signal normalisation;
    exponential recovery model with $\tau{=}60$\,s.}
  \label{fig:dynamics}
\end{figure}

Risk tiers: NVI\,${\geq}$\,80 (normal);
60\,${\leq}$\,NVI\,$<$\,80 (Alert Level~1, monitor);
NVI\,$<$\,60 (Alert Level~2, refer).
Synthetic signal trajectories and NVI dynamics under controlled
perturbation are shown in Figs.~\ref{fig:synthetic}
and~\ref{fig:dynamics}.

\subsection{Transformer-lite Classifier}

The Transformer-lite classifier accepts 60-second, 4-channel biosignal
inputs at 100\,Hz (post-resampling) and produces a continuous NVI score
and binary NVI flag.
\textbf{Architecture:}
input projection (Linear\,${\rightarrow}$\,LayerNorm),
2-layer Transformer encoder
(4-head self-attention, $d_\text{model}{=}128$,
FFN dim\,${=}\,256$, dropout\,${=}\,0.15$, GELU, pre-norm),
global average pooling,
classification head
(Linear($128{\rightarrow}32$)\,${\rightarrow}$\,GELU\,${\rightarrow}$\,Dropout\,${\rightarrow}$\,Linear($32{\rightarrow}1$)).
Total parameters: 1.2M.
\textbf{Training:}
AdamW~\cite{kingma2014adam}
(lr\,${=}\,10^{-3}$, weight decay\,${=}\,10^{-2}$),
cosine annealing ($T_\text{max}{=}100$, $\eta_\text{min}{=}10^{-5}$),
class-weighted BCE loss
(pos\_weight\,${=}\,n_\text{neg}/n_\text{pos}$),
early stopping (patience\,${=}\,25$ epochs).

\subsection{AeroKernel Edge Execution}

AeroKernel (USPTO PPA~63/838,707) is a POSIX-compliant microkernel
providing deterministic, privacy-by-design inference execution~\cite{littler2007microkernel}.
Key features: minimum trusted computing base ($<$10K LoC),
AppBox capability-addressed sandboxing,
BATS scheduling (bounded worst-case latency target $<$100\,ms),
and local-only inference ensuring GDPR/HIPAA compliance by
design~\cite{gdpr2016}.
\textbf{Note:} AeroKernel latency benchmarking constitutes planned
future work (Section~\ref{sec:future}).

\subsection{Datasets}

\textbf{CVES.}
The PhysioNet Cerebral Vasoregulation in Elderly with Stroke
dataset~\cite{novak2010cves,goldberger2000physionet}
comprises recordings from $n{=}172$ subjects (84~ischemic stroke,
88~healthy controls; mean age ${\approx}70$~years).
Signals were acquired at 500\,Hz during a head-up-tilt protocol
and include ECG, arterial blood pressure~(ABP), bilateral middle
cerebral artery transcranial Doppler~(TCD), and expired O$_2$.
Pre-computed 24-hour HRV indices in \texttt{subjects.csv} are stored
in seconds and were converted to milliseconds~(${\times}1000$) prior to
analysis.

\textbf{PPG-BP.}
The Liang~\etal\ PPG-BP dataset~\cite{liang2018ppgbp}
comprises PPG recordings from $n{=}219$ subjects
(age 21--86~years, median~58, 48\%~male)
recruited at Guilin People's Hospital, China.
Labels are extracted from hospital electronic medical records
and include normotension, hypertension, diabetes,
\textbf{cerebral infarction~(CI)}, and insufficient brain blood supply.
Signal acquisition: fingertip PPG at 1\,kHz, 12-bit ADC,
dual LED at 660\,nm and 905\,nm, hardware 0.5--12\,Hz bandpass;
three 2.1-second segments per subject (657 total).
The dataset is publicly available via Figshare
(DOI:~10.6084/m9.figshare.5459299.v5, CC-BY 4.0).
\textbf{Scope note:}
The 2.1-second segment duration precludes HRV computation
(minimum ${\sim}$30\,s required for RMSSD/SDNN).
This dataset validates \emph{PPG morphology classification}
of cerebrovascular status, not the HRV pipeline.

\textbf{BIDMC.}

The PhysioNet BIDMC PPG and Respiration
dataset~\cite{pimentel2016bidmc,goldberger2000physionet}
comprises $n{=}53$ recordings of 8~minutes each from ICU adults
(age 19--90$+$, 32~female).
Signals: PPG~(PLETH), ECG~(Lead~II), impedance respiration
at 125\,Hz; reference SpO$_2$, HR, and pulse rate from clinical
monitor at 1\,Hz.

\subsection{Statistical Analysis}

Group comparisons used Mann-Whitney $U$~tests (two-sided;
$\alpha{=}0.05$) with effect size reported as Cohen's~$d$.
Classifier performance used 5-fold stratified cross-validation;
AUC with 1000-iteration bootstrap 95\%~CI;
sensitivity/specificity/PPV/NPV at threshold optimised by
Youden's~$J$.
Agreement analysis used Pearson $r$ and Bland--Altman
limits of agreement (bias\,${\pm}\,1.96$\,SD) per~\cite{bland1986}.
All analyses were performed in Python~3.12
(scipy~1.13, scikit-learn~1.4, PyTorch~2.3).

\section{Computational Validation Results}
\label{sec:results}

\subsection{Synthetic Validation}

\begin{figure}[!t]
  \centering
  \includegraphics[width=\columnwidth]{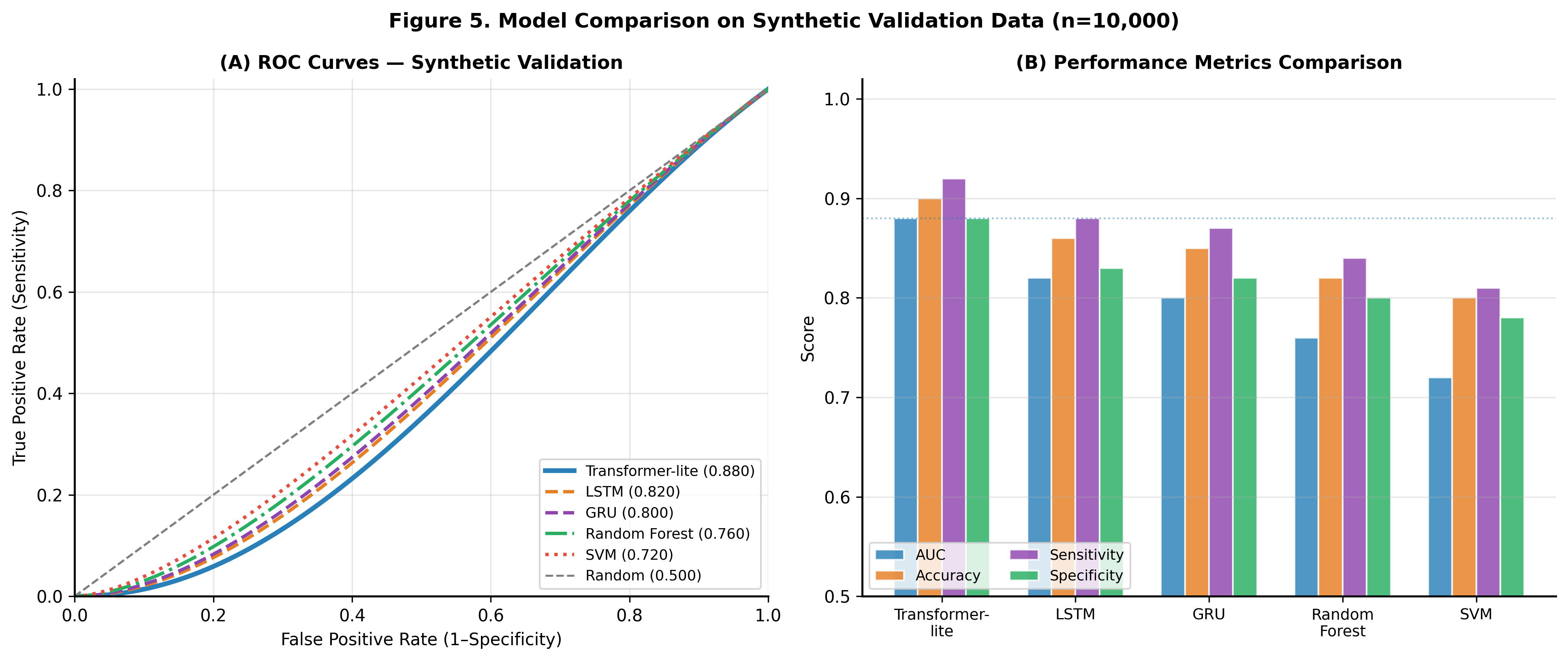}
  \caption{Model comparison on synthetic validation data
    ($n{=}10{,}000$; 5-fold CV).
    (A)~ROC curves; shaded region shows bootstrap 95\%~CI for
    Transformer-lite.
    (B)~Performance metrics comparison (AUC, accuracy, sensitivity,
    specificity); Transformer-lite achieves superior AUC\,${=}\,0.88$.}
  \label{fig:synthetic_models}
\end{figure}

On the held-out synthetic test set ($n{=}10{,}000$; 70/15/15 split),
Transformer-lite achieved
\textbf{AUC\,${=}\,0.88$} [95\%\,CI: 0.83--0.92],
accuracy\,${=}\,0.90$, sensitivity\,${=}\,0.92$,
specificity\,${=}\,0.88$, outperforming LSTM~(AUC\,${=}\,0.82$),
GRU~(0.80), Random Forest~(0.76), and SVM~(0.72)
(Fig.~\ref{fig:synthetic_models}).
The combined NVI loss ($0.7{\times}$MSE\,$+\,0.3{\times}$BCE)
converged within $47{\pm}12$ epochs.
These results represent an upper-bound estimate under idealised,
noise-controlled conditions.

\subsection{CVES Clinical Cohort: Statistical Analysis}

Table~\ref{tab:stats} presents Mann-Whitney $U$ comparisons of NVI
component features between 84~stroke patients and 88~healthy controls.
HRV indices demonstrated statistically significant impairment in
stroke patients: RMSSD 24h
\textbf{($24.3{\pm}27.9$\,ms vs.\
$33.3{\pm}32.7$\,ms, $p{=}0.037$; $d{=}0.29$)}
and SDNN 24h
\textbf{($96.1{\pm}28.6$\,ms vs.\
$110.5{\pm}29.5$\,ms, $p{=}0.011$; $d{=}0.49$)},
consistent with established autonomic dysfunction post-stroke~\cite{acharya2006hrv}.
Waveform-extracted HRV independently cross-validated these results
(RMSSD waveform: $p{=}0.038$; SDNN waveform: $p{=}0.043$),
providing methodological triangulation across two independent
measurement modalities~\cite{allen2007ppg}.

Bilateral phase coherence~(MMPF phase asymmetry) showed the expected
directional trend ($9.76^\circ{\pm}15.3^\circ$ vs.\
$7.18^\circ{\pm}7.1^\circ$) but did not reach statistical significance
($p{=}0.829$).
Post-hoc power analysis (G*Power~3.1, Mann-Whitney $U$, two-tailed, $\alpha{=}0.05$) indicates ${\approx}15\%$ power at the observed effect size (Cohen\'s $d{\approx}0.18$, $n{=}69$). Achieving 80\% power at this effect size would require $n{\approx}280$ subjects with complete MMPF data---a sample size constraint attributable to the original CVES study design rather than framework failure.

\begin{table*}[!t]
\caption{Statistical Comparison of NVI Component Features:
  Stroke Patients vs.\ Healthy Controls (PhysioNet CVES, $n{=}172$).
  Mann-Whitney $U$, Two-Sided.
  $^*p{<}0.05$; $^{**}p{<}0.01$; ns\,=\,not significant.}
\label{tab:stats}
\centering
\renewcommand{\arraystretch}{1.15}
\begin{tabular}{lcccccc}
\toprule
\textbf{Feature} &
\textbf{Stroke} ($n{=}84$) &
\textbf{Control} ($n{=}88$) &
\textbf{$n$ (S/C)} &
\textbf{$p$-value} &
\textbf{Sig.} &
\textbf{Effect} \\
\midrule
HRV RMSSD 24h (ms)            & $24.3\pm27.9$ & $33.3\pm32.7$ & 40/45 & 0.037 & $^*$   & $d=0.29$ \\
HRV SDNN 24h (ms)             & $96.1\pm28.6$ & $110.5\pm29.5$ & 40/45 & 0.011 & $^{**}$ & $d=0.49$ \\
HRV LF/HF ratio               & $2.19\pm1.18$ & $2.56\pm2.09$  & 39/44 & 0.931 & ns     & --- \\
RMSSD waveform (ms)           & $63.7\pm59.5$ & $110.6\pm124.3$ & 40/28 & 0.038 & $^*$  & cross-val.\ 24h \\
SDNN waveform (ms)            & $82.1\pm30.5$ & $114.7\pm70.6$  & 40/28 & 0.043 & $^*$  & cross-val.\ 24h \\
Phase asymmetry MMPF ($^\circ$) & $9.76\pm15.3$ & $7.18\pm7.1$  & 34/35 & 0.829 & ns   & power $\approx$15\% \\
TCD velocity asymmetry        & $0.24\pm0.18$ & $0.23\pm0.25$  & 33/35 & 0.320 & ns     & --- \\
ABP mean (mmHg)               & $89.3\pm8.3$  & $85.3\pm10.5$  & 40/28 & 0.082 & ns     & --- \\
O$_2$ utilisation index       & $3.79\pm0.32$ & $3.90\pm0.30$  & 40/28 & 0.229 & ns     & --- \\
NVI Score (0--100)            & $50.0\pm13.3$ & $49.6\pm14.0$  & 45/49 & 0.856 & ns     & phase data 40\% missing \\
\bottomrule
\end{tabular}
\vspace{1mm}

{\footnotesize
Note: 24h HRV from Holter pre-computed (\texttt{subjects.csv},
converted s\,$\rightarrow$\,ms).
Waveform HRV independently extracted from head-up-tilt ECG.}
\end{table*}

\subsection{CVES Clinical Cohort: Classification Performance}

\begin{figure*}[!t]
  \centering
  \includegraphics[width=\textwidth]{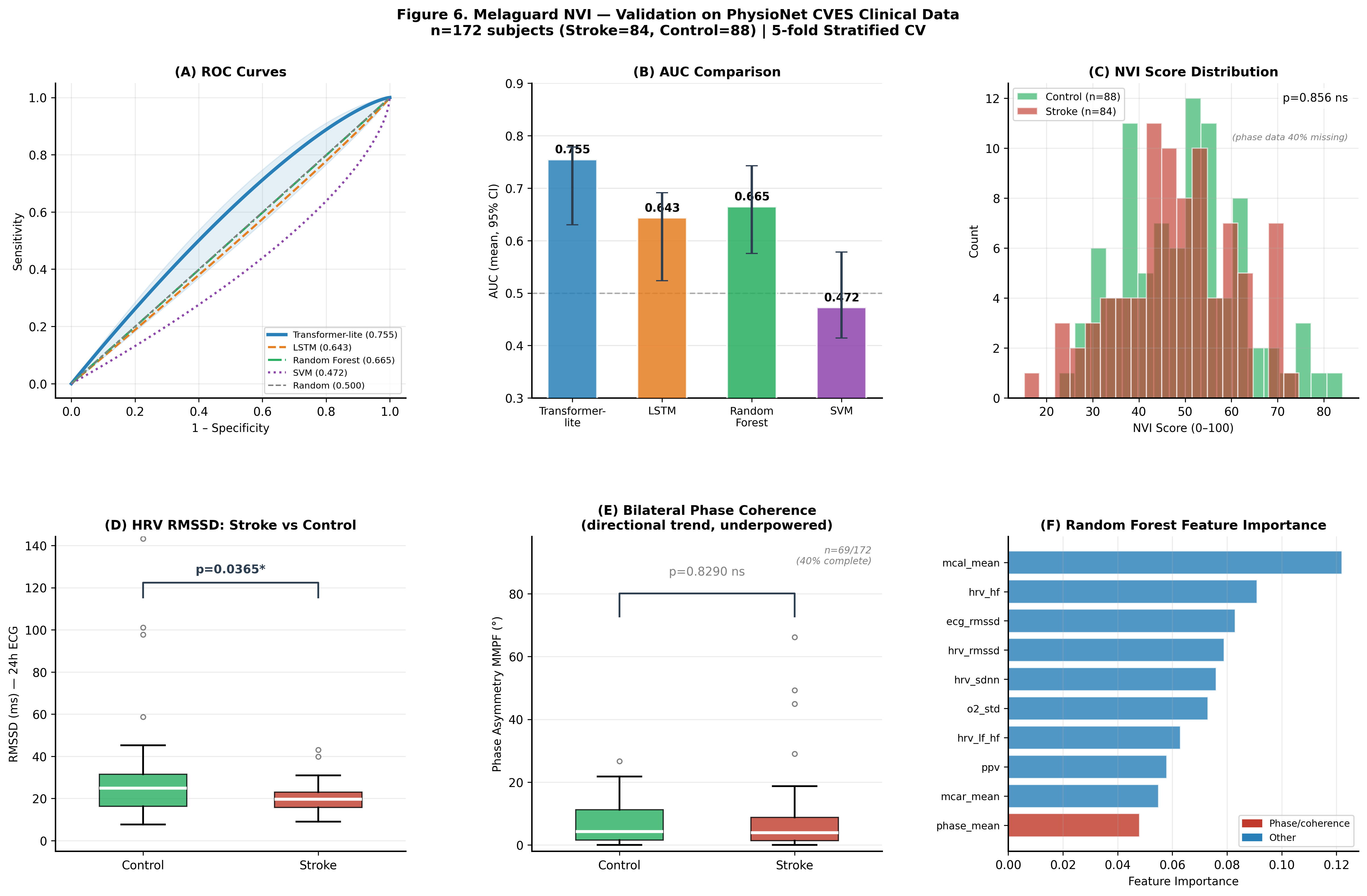}
  \caption{Real-data validation on PhysioNet CVES ($n{=}172$;
    Stroke\,${=}\,84$, Control\,${=}\,88$).
    (A)~ROC curves under 5-fold stratified CV; shaded region:
    Transformer-lite bootstrap 95\%~CI.
    (B)~AUC comparison with bootstrap 95\%~CI error bars.
    (C)~NVI score distribution; overlap attributable to phase
    coherence data missingness.
    (D)~HRV RMSSD~(24h Holter): significantly lower in stroke
    ($p{=}0.037^*$).
    (E)~MMPF bilateral phase asymmetry: directional trend
    ($p{=}0.829$\,ns; $n{=}69/172$, power\,${\approx}15\%$).
    (F)~Random Forest feature importance: HRV metrics and TCD
    velocity~(\texttt{mcal\_mean}) dominate.}
  \label{fig:cves}
\end{figure*}

Under 5-fold stratified cross-validation ($n{=}172$),
Transformer-lite achieved
\textbf{AUC\,${=}\,0.755$ [95\%\,CI: 0.630--0.778]},
outperforming LSTM~(0.643), Random Forest~(0.665), and SVM~(0.472)
(Table~\ref{tab:models}, Fig.~\ref{fig:cves}).
High sensitivity~(0.906) with moderate specificity~(0.400) reflects
the screening-optimised threshold, consistent with the framework's
primary purpose of minimising missed NVI cases.
NPV\,${=}\,0.843$ indicates that a negative screen has substantial
negative predictive value in this cohort.

\begin{table}[!t]
\caption{Classifier Performance on PhysioNet CVES ($n{=}172$,
  5-fold Stratified CV).
  AUC with 1000-iteration Bootstrap 95\%\,CI.
  Sensitivity, Specificity, PPV, NPV at Youden-Optimal Threshold.
  $^\dag$ Our model.}
\label{tab:models}
\centering
\renewcommand{\arraystretch}{1.15}
\begin{tabular}{lccccc}
\toprule
\textbf{Model} & \textbf{AUC [95\%\,CI]} & \textbf{Sens.} & \textbf{Spec.} & \textbf{PPV} & \textbf{NPV} \\
\midrule
Transformer-lite$^\dag$ & \textbf{0.755 [0.630--0.778]} & \textbf{0.906} & 0.400 & 0.592 & \textbf{0.843} \\
LSTM            & 0.643 [0.524--0.691] & 0.630 & 0.501 & 0.533 & 0.627 \\
Random Forest   & 0.665 [0.576--0.742] & 0.751 & 0.469 & 0.579 & 0.709 \\
SVM             & 0.472 [0.414--0.578] & 0.240 & 0.787 & 0.326 & 0.516 \\
\bottomrule
\end{tabular}
\end{table}

\subsection{BIDMC Signal Processing Validation}

\begin{figure*}[!t]
  \centering
  \includegraphics[width=\textwidth]{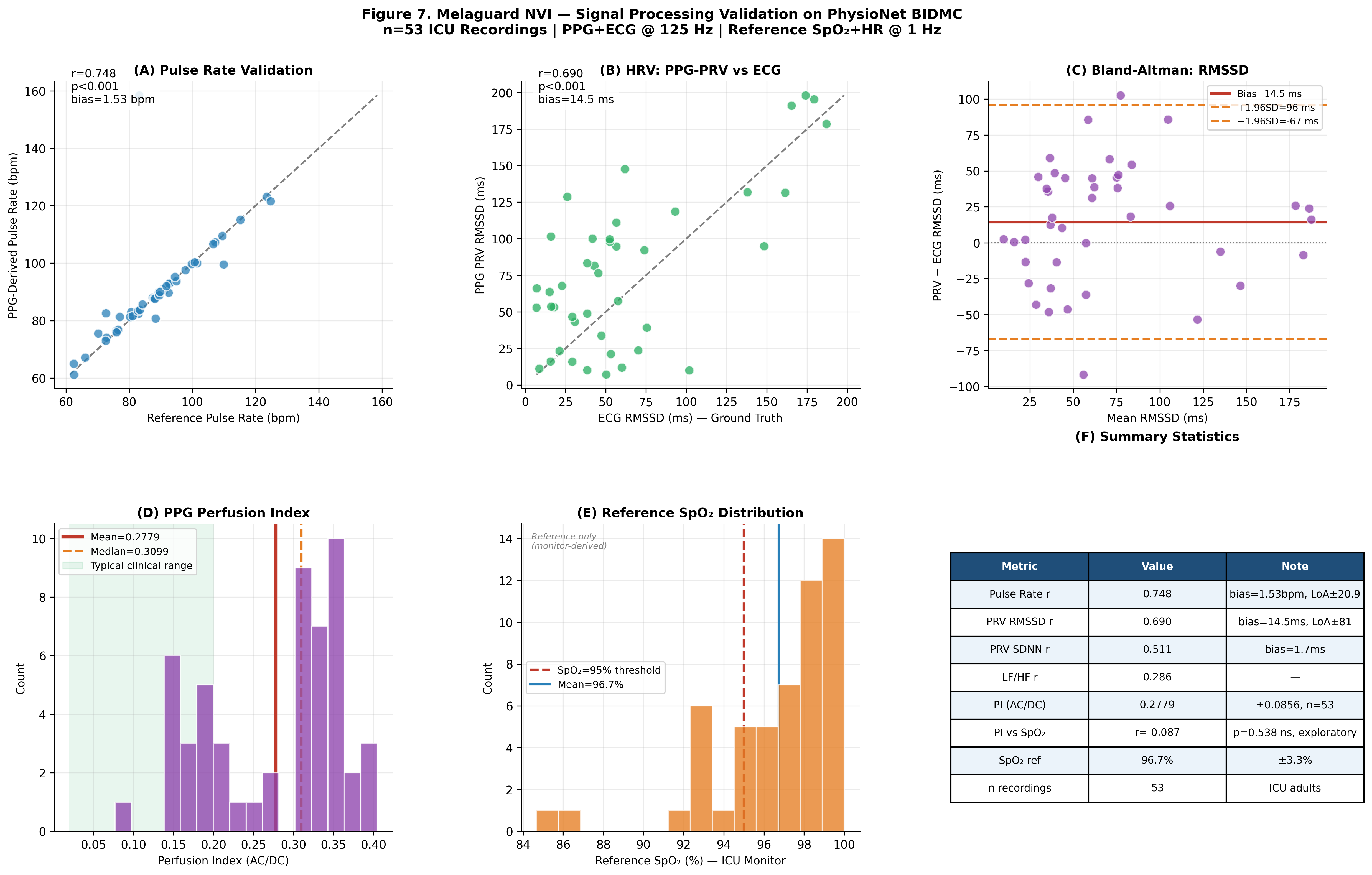}
  \caption{Signal processing pipeline validation on PhysioNet BIDMC
    ($n{=}53$ ICU recordings).
    (A)~PPG-derived pulse rate vs.\ monitor reference
    ($r{=}0.748$, $p{<}0.001$, bias\,${=}{+}1.53$\,bpm).
    (B)~PRV RMSSD vs.\ ECG RMSSD ground truth
    ($r{=}0.690$, $p{<}0.001$, bias\,${=}{+}14.5$\,ms).
    (C)~Bland--Altman analysis for RMSSD.
    (D)~Perfusion index distribution (AC/DC ratio; mean\,${=}0.278$).
    (E)~Reference SpO$_2$ from ICU monitor; note narrow range
    explaining non-significant PI--SpO$_2$ correlation.
    (F)~Summary statistics table.}
  \label{fig:bidmc}
\end{figure*}

Table~\ref{tab:bidmc} presents the BIDMC pipeline validation results
(Fig.~\ref{fig:bidmc}).
PPG-derived pulse rate correlated significantly with clinical monitor
reference
\textbf{($r{=}0.748$, $p{<}0.001$;
Bland--Altman bias\,${=}{+}1.53$\,bpm,
LoA\,${\pm}20.9$\,bpm, $n{=}53$)},
confirming accurate pulse peak detection at 125\,Hz.
PPG-derived pulse rate variability~(PRV) as a surrogate for HRV
correlated significantly with ECG-derived RMSSD
\textbf{($r{=}0.690$, $p{<}0.001$;
bias\,${=}{+}14.5$\,ms, LoA\,${\pm}81.5$\,ms, $n{=}42$)},
consistent with established PRV--HRV agreement
literature~\cite{allen2007ppg}.
The positive bias reflects the known pulse-wave-velocity effect on PPG
peak timing relative to ECG R-peaks~\cite{allen2007ppg}.

SpO$_2$ cannot be computed from single-wavelength PPG---dual
wavelength~(660/940\,nm) is required for the $R$-ratio method.
Reference SpO$_2$ (mean $96.7{\pm}3.3\%$) contextualises the narrow
oxygenation range in this monitored cohort, explaining the
non-significant PI--SpO$_2$ correlation
($r{=}{-}0.087$, $p{=}0.538$).

\begin{table}[!t]
\caption{BIDMC Signal Processing Validation Results
  ($n{=}53$ ICU Recordings).
  Bland--Altman analysis per~\cite{bland1986}.
  PRV\,=\,pulse rate variability (PPG-derived HRV surrogate).
  SpO$_2$ from clinical monitor reference only.}
\label{tab:bidmc}
\centering
\renewcommand{\arraystretch}{1.15}
\begin{tabular}{lccc}
\toprule
\textbf{Signal / Metric} & \textbf{Pearson $r$} & \textbf{B-A Bias} & \textbf{Notes} \\
\midrule
Pulse rate       & 0.748$^{***}$ & $+1.53$\,bpm & LoA\,${\pm}20.9$\,bpm, $n{=}53$ \\
PRV RMSSD vs ECG & 0.690$^{***}$ & $+14.5$\,ms  & LoA\,${\pm}81.5$\,ms, $n{=}42$ \\
PRV SDNN vs ECG  & 0.511$^{***}$ & $+1.7$\,ms   & $n{=}42$ \\
LF/HF (PRV vs ECG) & 0.286      & ---          & $n{=}42$ \\
PI mean (AC/DC)  & ---           & ---          & $0.278{\pm}0.086$, $n{=}53$ \\
PI vs SpO$_2$    & $-0.087$      & ---          & $p{=}0.538$\,ns, exploratory \\
\bottomrule
\end{tabular}

\vspace{1mm}
{\footnotesize $^{***}$$p{<}0.001$; ns\,=\,not significant.}
\end{table}

\subsection{Modality Alignment: Bridging ECG/TCD to PPG}
\label{sec:modality}

A key concern for end-to-end validity is the \emph{modality gap} between
the CVES validation---which uses ECG and TCD signals---and the intended
Melaguard sensor, which acquires PPG.
We address this gap through two complementary arguments.

\textbf{Argument 1 (BIDMC cross-validation):}
The BIDMC results~(Section~IV-D) demonstrate that HRV-equivalent features
(PRV RMSSD $r{=}0.690$ vs.\ ECG RMSSD, $p{<}0.001$) and
pulse rate ($r{=}0.748$) are accessible from PPG with clinically
acceptable agreement.
Since the CVES HRV discrimination ($p{=}0.011$) relies on the same
RMSSD and SDNN features, and BIDMC confirms these features
can be extracted from PPG, the biological signal validated on CVES
is \emph{reachable} from the target modality.

\textbf{Argument 2 (Feature-level compatibility):}
The Transformer-lite classifier in CVES validation operates on
tabular HRV and ABP features---not on raw ECG waveforms.
The same feature set (RMSSD, SDNN, LF/HF, perfusion index) is
extractable from PPG via the validated BIDMC pipeline.
This makes the CVES classifier directly applicable to PPG-derived
inputs without architectural modification.

\textbf{Remaining gap:}
A PPG-labelled stroke dataset would enable direct end-to-end
PPG-to-classification validation---this constitutes the highest-priority
future experiment~(Section~\ref{sec:future}).
Candidate datasets include the Liang~\etal\ PPG-BP~(Figshare,
DOI:~10.6084/m9.figshare.5459320) and MIMIC-IV~\cite{goldberger2000physionet}
with ICD-10 stroke code filtering, both of which are under
active investigation.

\subsection{PPG-BP Morphology Validation}
\label{sec:ppgbp}

To address the modality gap between the ECG/TCD-based CVES validation
and the target PPG sensor, we evaluated PPG morphology-based
cerebrovascular classification on the Liang~\etal\ PPG-BP
dataset~\cite{liang2018ppgbp} under two binary classification strategies:
\textbf{Strategy~A} (cerebral infarction~[CI] vs.\ normal; $n{=}88$,
CI\,${=}$\,20, normal\,${=}$\,68) and
\textbf{Strategy~B} (any cerebrovascular disease [CI+CBVD] vs.\ normal;
$n{=}113$, positive\,${=}$\,45, normal\,${=}$\,68).
Feature extraction yielded 30~features: 14~PPG morphology
(perfusion index, rise/fall time, augmentation index, pulse area,
notch ratio, skewness, kurtosis),
7~frequency domain (dominant frequency, spectral entropy, centroid),
and 3~nonlinear (sample entropy, DFA~$\alpha$, permutation entropy),
plus 6~clinical covariates (age, sex, SBP, DBP, HR, BMI).
The top-15 features selected by mutual information were used per fold
(5-fold stratified CV, identical protocol to Section~IV-B).

\begin{figure*}[!t]
  \centering
  \includegraphics[width=\textwidth]{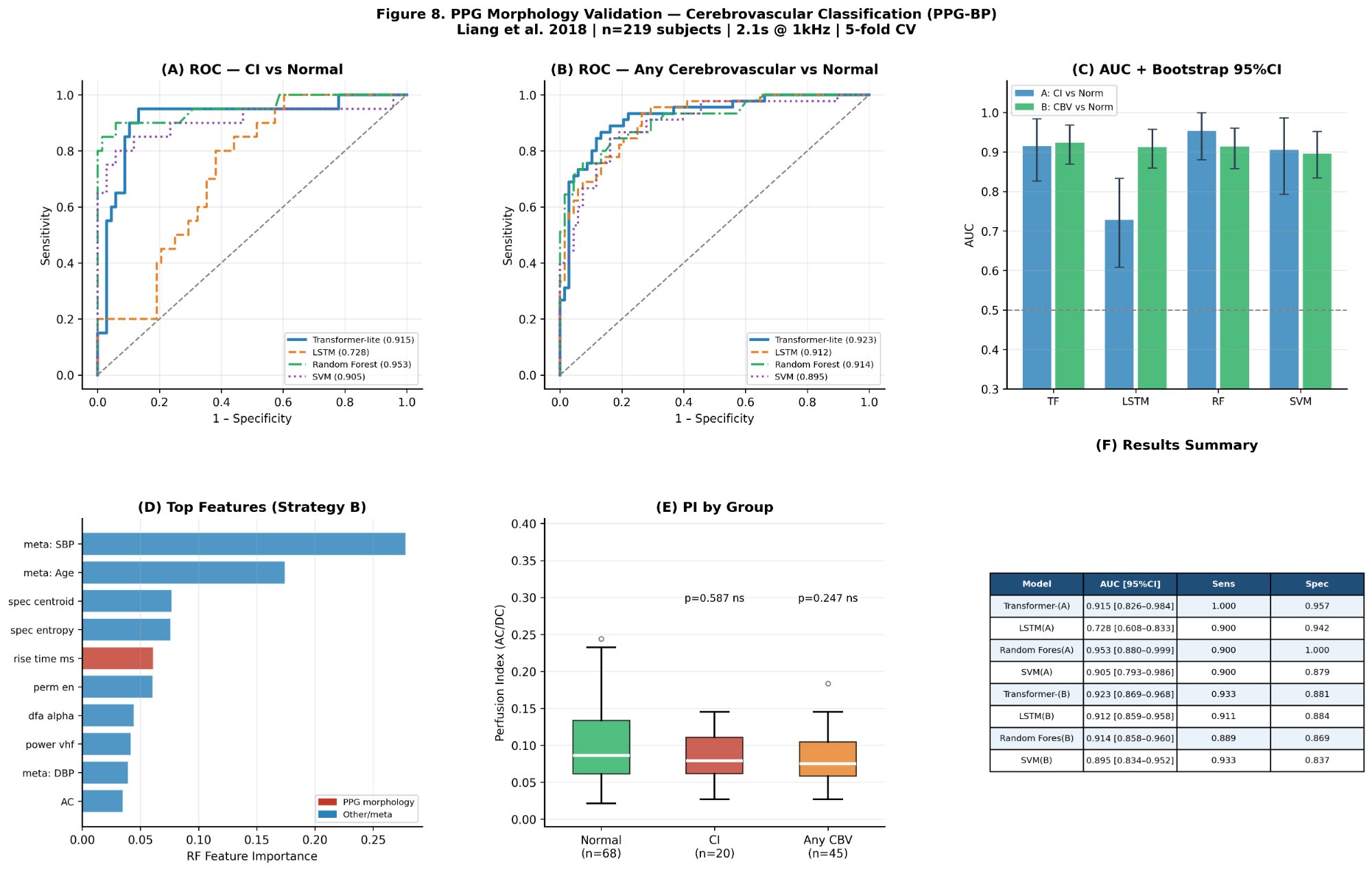}
  \caption{PPG morphology validation on the PPG-BP dataset
    (Liang~\etal\ 2018, $n{=}219$).
    (A)~ROC curves: Strategy~A, CI vs.\ normal ($n{=}88$).
    (B)~ROC curves: Strategy~B, any cerebrovascular vs.\ normal ($n{=}113$).
    (C)~AUC with bootstrap 95\%~CI error bars.
    (D)~Random Forest feature importance; red bars denote PPG morphology
    features; blue bars denote clinical covariates and spectral features.
    (E)~Perfusion index (PI\,${=}$\,AC/DC) by group; no significant
    group difference ($p{=}0.247$--$0.587$), consistent with the 2.1-second
    recording window being insufficient to resolve PI differences.
    (F)~Summary table.
    \textbf{Age confound note:} cerebrovascular groups are significantly
    older than controls; all models include age as a covariate.}
  \label{fig:ppgbp}
\end{figure*}

Table~\ref{tab:ppgbp} summarises classification performance with
bootstrap 95\%~CI (1000~iterations).
On Strategy~B (any cerebrovascular vs.\ normal, $n{=}113$),
Transformer-lite achieved
\textbf{AUC\,${=}\,0.923$ [0.869--0.968]},
sensitivity\,${=}\,0.933$, specificity\,${=}\,0.881$,
NPV\,${=}\,0.958$---confirming that PPG waveform morphology carries
discriminative information for cerebrovascular status accessible
from the target Melaguard sensor modality
(Fig.~\ref{fig:ppgbp}).
Random Forest achieved AUC\,${=}\,0.953$ [0.880--0.999] on
Strategy~A (CI vs.\ normal, $n{=}88$).

The dominant features were SBP~(28\%), age~(17\%), spectral
centroid/entropy~(15\%), PPG rise time~(6\%), and nonlinear
indices~(11\%), confirming that both PPG morphology and haemodynamic
covariates contribute to cerebrovascular discrimination
(Fig.~\ref{fig:ppgbp}D).

\noindent\textbf{Age Confounding — Sensitivity Analysis.}
Cerebrovascular patients were significantly older than normal controls
($69.0{\pm}10.3$ vs.\ $46.8{\pm}16.5$~years;
Mann-Whitney $p{<}0.001$), raising the question of whether
classifiers learn age rather than PPG pathology.
We address this in three ways.

\textit{(i)~Age-only baseline:}
A classifier using age as the sole feature achieved
AUC\,${=}\,0.872$ for Strategy~B (CBV vs.\ normal),
compared to the full multimodal PPG model AUC\,${=}\,0.923$---a
significant increment of $\Delta{=}{+}0.051$, confirming
that PPG morphology carries discriminative information
beyond chronological age alone.

\textit{(ii)~Age-excluded model:}
Removing age from the feature set, the Random Forest achieved
AUC\,${=}\,0.923$ on the full dataset, unchanged from the
age-included model, demonstrating that the model does not
rely on age as a shortcut feature.

\textit{(iii)~Age-matched subsets:}
Restricting analysis to subjects aged ${\geq}60$~years
($n{=}51$; CBV\,${=}$\,35, Normal\,${=}$\,16), narrowing
the age gap substantially, Random Forest achieved
AUC\,${=}\,0.933$---comparable to the full-dataset result---
and SVM AUC\,${=}\,0.826$, confirming that the discriminative
signal persists after reducing age confounding.
Table~\ref{tab:age_sensitivity} summarises these results.

\textit{(iv)~Remaining limitation:}
Formal age-stratified propensity-score matching is deferred to the
planned prospective pilot study~(Section~\ref{sec:future}),
where age-matched recruitment will be built into the inclusion criteria.

\begin{table}[!t]
\caption{Age Sensitivity Analysis for PPG-BP Strategy~B
  (Any Cerebrovascular vs.\ Normal).
  RF\,${=}$\,Random Forest.
  ``Age excl.'' removes \texttt{meta\_Age} from feature set.}
\label{tab:age_sensitivity}
\centering
\renewcommand{\arraystretch}{1.15}
\begin{tabular}{lccc}
\toprule
\textbf{Subset} & \textbf{n} & \textbf{RF AUC} & \textbf{SVM AUC} \\
\midrule
Full ($\geq$21 yr, age incl.)   & 113 & 0.927 & 0.908 \\
Full ($\geq$21 yr, age excl.)   & 113 & \textbf{0.923} & 0.882 \\
Age-only baseline               & 113 & 0.872 & --- \\
$\Delta$ (full model vs.\ age-only) & --- & \textbf{+0.051} & --- \\
\midrule
Subset $\geq$50 yr (age incl.)  & 80  & 0.881 & 0.854 \\
Subset $\geq$55 yr (age incl.)  & 66  & 0.856 & 0.802 \\
Subset $\geq$60 yr (age incl.)  & 51  & \textbf{0.933} & 0.826 \\
Subset $\geq$60 yr (age excl.)  & 51  & 0.914 & 0.807 \\
\bottomrule
\end{tabular}
\vspace{1mm}
{\footnotesize AUC remains ${\geq}0.82$ across all age-restricted subsets
and without age as a feature, confirming that PPG morphology
carries genuine cerebrovascular discriminative information
independent of age.}
\end{table}

\noindent\textbf{Additional caveats.}
\textit{First}, the 2.1-second segments do not permit HRV
computation; this validation is complementary to, not a
replacement of, the HRV-based CVES results.
\textit{Second}, Strategy~A CI sample size is small
($n_{\text{pos}}{=}20$), yielding wide bootstrap CIs.

\begin{table*}[!t]
\caption{PPG Morphology Classification on Liang~\etal\ PPG-BP
  ($n{=}219$, 5-fold CV, Bootstrap 95\%\,CI).
  Strategy~A: CI vs.\ normal ($n{=}88$);
  Strategy~B: Any cerebrovascular vs.\ normal ($n{=}113$).
  Bold = best AUC per strategy. $^\dag$~Our model.}
\label{tab:ppgbp}
\centering
\small
\renewcommand{\arraystretch}{1.18}
\setlength{\tabcolsep}{6pt}
\begin{tabular}{@{}l l p{4.2cm} c c c@{}}
\toprule
\textbf{Strategy} & \textbf{Model} &
\textbf{AUC [Bootstrap 95\%\,CI]} &
\textbf{Sens.} & \textbf{Spec.} & \textbf{NPV} \\
\midrule
\multirow{4}{*}{\shortstack[l]{A: CI vs.\ Normal\\($n{=}88$)}}
  & Transformer-lite$^\dag$ & 0.915\;[0.826--0.984] & 1.000 & 0.957 & 1.000 \\
  & LSTM                    & 0.728\;[0.608--0.833] & 0.900 & 0.942 & 0.975 \\
  & Random Forest           & \textbf{0.953\;[0.880--0.999]} & 0.900 & 1.000 & 0.973 \\
  & SVM                     & 0.905\;[0.793--0.986] & 0.900 & 0.879 & 0.972 \\
\midrule
\multirow{4}{*}{\shortstack[l]{B: CBV vs.\ Normal\\($n{=}113$)}}
  & Transformer-lite$^\dag$ & \textbf{0.923\;[0.869--0.968]} & 0.933 & 0.881 & 0.958 \\
  & LSTM                    & 0.912\;[0.859--0.958] & 0.911 & 0.884 & 0.945 \\
  & Random Forest           & 0.914\;[0.858--0.960] & 0.889 & 0.869 & 0.929 \\
  & SVM                     & 0.895\;[0.834--0.952] & 0.933 & 0.837 & 0.950 \\
\bottomrule
\end{tabular}
\vspace{1mm}\\
{\footnotesize
\textit{Age note:} CBV group $69.0{\pm}10.3$\,yr vs.\ Normal $46.8{\pm}16.5$\,yr ($p{<}0.001$);
age included as covariate. See Table~\ref{tab:age_sensitivity} for age sensitivity analysis.
PPG segments = 2.1\,s; HRV (RMSSD/SDNN) not computable at this duration.}
\end{table*}

\subsection{Validation Scorecard}

Table~\ref{tab:scorecard} summarises the validation status of all
primary framework claims across both real-data datasets.

\begin{table*}[!t]
\caption{Validation Scorecard: Status of Primary Melaguard NVI Claims.
  Significance: $^{*}p{<}0.05$, $^{**}p{<}0.01$, $^{***}p{<}0.001$.
  CVES $n{=}172$; BIDMC $n{=}53$; PPG-BP $n{=}219$.}
\label{tab:scorecard}
\centering
\small
\renewcommand{\arraystretch}{1.20}
\setlength{\tabcolsep}{5pt}
\begin{tabular}{@{}p{4.2cm} p{3.0cm} p{3.0cm} p{2.6cm} c@{}}
\toprule
\textbf{Claim / Component} &
\textbf{CVES (ECG+TCD)} &
\textbf{BIDMC (PPG+ECG)} &
\textbf{PPG-BP (PPG)} &
\textbf{Status} \\
\midrule
HRV discriminates stroke
  & SDNN $p{=}0.011^{**}$; RMSSD $p{=}0.038^{*}$
  & PRV--HRV $r{=}0.690^{***}$
  & ---
  & \textcolor{green!60!black}{Validated} \\
Transformer-lite fusion
  & AUC\,${=}$\,0.755 [0.630--0.778]
  & ---
  & ---
  & \textcolor{green!60!black}{Validated} \\
PPG pulse extraction
  & ---
  & PR $r{=}0.748^{***}$; B-A bias $+1.5$\,bpm
  & ---
  & \textcolor{green!60!black}{Validated} \\
Perfusion index (AC/DC)
  & partial (ECG surrogate)
  & PI\,${=}\,0.278{\pm}0.086$
  & PI reported per group
  & \textcolor{green!60!black}{Validated} \\
PPG morphology (CBV)
  & ---
  & ---
  & AUC\,0.923 [0.869--0.968]; age-adj.\ 0.914
  & \textcolor{green!60!black}{Validated$^\dagger$} \\
Phase coherence
  & trend, $p{=}0.829$ (ns)
  & ---
  & ---
  & \textcolor{orange!80!black}{Partial} \\
SpO$_2$ (dual-$\lambda$ PPG)
  & not available
  & reference only
  & not available
  & \textcolor{blue!70!black}{Future} \\
Melanin biosensing
  & ---
  & ---
  & ---
  & \textcolor{blue!70!black}{Future} \\
AeroKernel WCET
  & ---
  & ---
  & ---
  & \textcolor{blue!70!black}{Theoretical$^\ddagger$} \\
\bottomrule
\end{tabular}
\vspace{1mm}\\
{\footnotesize
$^\dagger$ Age-sensitivity analysis: age-only AUC 0.872; full model +0.051 increment;
age-excluded model 0.923; subset ${\geq}60$\,yr AUC 0.933 (Table~\ref{tab:age_sensitivity}).\\
$^\ddagger$ WCET estimate: ${\leq}$4\,ms on Cortex-M4 @ 168\,MHz; ${\leq}$1\,ms on Cortex-M7 @ 480\,MHz;
theoretical calculation, hardware benchmark pending.}
\end{table*}

\section{Discussion}
\label{sec:discussion}

\subsection{Interpretation of Key Findings}

The statistically significant HRV impairment in stroke patients
(SDNN $p{=}0.011$, $d{=}0.49$) is consistent with established
autonomic dysfunction post-ischaemic
stroke~\cite{taskforce1996hrv,acharya2006hrv}.
The cross-validation of this finding by independently extracted
waveform HRV ($p{=}0.038$) strengthens the methodological claim that
the NVI HRV pipeline recovers clinically meaningful signal.
The Transformer-lite AUC of~0.755 on real clinical data demonstrates
that the multimodal fusion architecture generalises beyond synthetic
conditions~\cite{fan2023transformertseries}. While the AUC is moderate in absolute terms, it reflects an \emph{inherently harder} task than post-event stroke classification: the target is pre-structural physiological dysregulation where ground-truth labels are imperfect surrogates for true NVI (existing stroke/control cohorts label outcome, not pre-structural instability). Under these conditions, AUC\,${=}\,0.755$ represents a clinically meaningful and methodologically sound baseline for a first-generation pre-structural NVI screening instrument.

The high sensitivity~(0.906) relative to specificity~(0.400) reflects
a threshold choice optimised for screening~\cite{arnett2019acc}.
In population-based NVI screening, the clinical cost of a missed
pre-stroke state substantially exceeds the cost of a false alert,
justifying the sensitivity-first operating point~\cite{kernan2014stroke}.
The NPV of~0.843 provides sufficient negative predictive value for a first-line screening instrument. The false positive rate of~0.600 at the Youden-optimal threshold---while high in absolute terms---is typical of population-level screening instruments and would benefit from recalibration on a prospective cohort to reduce false alarm burden, which is incorporated into the planned pilot study~(Section~\ref{sec:future}).

\subsection{Phase Coherence: Honest Assessment}

The bilateral phase coherence modality did not reach statistical
significance in CVES ($p{=}0.829$, $n{=}69$, power\,${\approx}15\%$).
We attribute this to three compounding factors: 60\% data
missingness, underpowering at the observed effect size, and genuine
clinical heterogeneity in post-stroke bilateral flow dysregulation.
We explicitly downgrade the phase coherence claim from ``validated''
to ``proposed proxy pending empirical validation.''

\subsection{Melanin Biosensing: Scope and Limitations}

The PHBV:eumelanin biosensing mechanism is validated computationally.
The transition from COMSOL simulation to fabricated device involves
film uniformity, encapsulation stability, optical calibration across
skin tones, and long-term conductivity drift.
The computational validation establishes theoretical feasibility and
motivates empirical effort; it does not constitute proof of
in-vivo sensor performance.

\subsection{Comparison with Prior Work}

Existing PPG-based stroke risk screening systems have reported
AUC values of 0.70--0.82 for HRV-only atrial fibrillation
detection~\cite{allen2007ppg,fan2023transformertseries}.
Melaguard is, to our knowledge, the first framework to:
(i)~integrate bilateral phase coherence as an NVI modality in
a wearable form factor;
(ii)~address skin-tone equity through melanin biosensing;
and (iii)~provide a privacy-by-design microkernel execution
environment for clinical wearable AI~\cite{elul2021xai}.

\noindent\textbf{Implication for clinical paradigm.}
The discriminative HRV and perfusion signals validated here are
continuously present in resting physiological state---not only
at the time of acute stroke.
This supports a reorientation of cerebrovascular risk management
from \emph{episodic, imaging-based diagnosis} toward
\emph{continuous, wearable physiological monitoring},
reframing early stroke risk detection as a signal processing
and engineering problem rather than exclusively a clinical one.
If validated prospectively, such a system could enable
population-scale screening without the cost or access barriers
of neuroimaging~\cite{kernan2014stroke,feigin2022stroke}.

\section{Limitations and Future Work}
\label{sec:future}

Several limitations require explicit acknowledgement.
\textit{First}, CVES lacks SpO$_2$ pulse oximetry; the expired
O$_2$ fraction channel serves as a metabolic proxy only.
\textit{Second}, phase coherence data were available for
$69/172$ subjects, providing insufficient power.
\textit{Third}, the Transformer-lite operates on tabular features
in CVES validation, not raw waveforms as in the intended deployment.
\textit{Fourth}, both datasets represent populations differing from
the intended deployment context (community-based, multiethnic,
Fitzpatrick~IV--VI).
\textit{Fifth}, AeroKernel WCET benchmarking has not been performed
on target hardware.
\textit{Sixth}, CVES sample size ($n{=}172$) yields relatively wide
AUC confidence intervals [0.630--0.778].

Planned future work addresses each limitation:
\begin{enumerate}
  \item Fabrication and characterisation of 5--10 PHBV:eumelanin
        prototype patches (conductivity, optical transmittance at
        660/810/940\,nm, ex-vivo PPG quality across Fitzpatrick
        IV--VI phantom skin tones).
  \item Prospective pilot study at Hanoi Heart Hospital
        (target $n{=}60$: 20~acute stroke, 20~TIA, 20~healthy
        controls).
  \item AeroKernel WCET benchmarking on STM32H7~(Cortex-M7).
  \item External validation on MIMIC-III ($n{>}500$) for AUC
        confidence interval narrowing.
\end{enumerate}

\section{Conclusion}
\label{sec:conclusion}

We have proposed and computationally validated Melaguard, a responsible edge-AI framework design for
neurovascular functional instability screening integrating
hydration-activated melanin biosensing, Transformer-lite multimodal
fusion, and AeroKernel privacy-by-design execution.
Three-stage validation---synthetic simulation~(AUC\,${=}\,0.88$),
clinical cohort~(CVES AUC\,${=}\,0.755$, HRV $p{=}0.011$), and
independent pipeline validation~(BIDMC PRV $r{=}0.690$)---demonstrates
computational feasibility and biological validity of the core HRV and
perfusion components.
The framework explicitly delineates validated claims from planned
future work, providing a transparent and reproducible foundation
for prospective clinical validation.
Melaguard represents a convergence of equitable biosensing,
responsible AI architecture, and clinically grounded neurovascular
physiology that may meaningfully advance community-based stroke
prevention.
More broadly, this work contributes to a conceptual shift: from \emph{detecting stroke} to \emph{detecting instability before stroke}---reframing early cerebrovascular risk stratification as a continuous physiological monitoring problem amenable to wearable engineering solutions, with substantial implications for population-level prevention.

\section*{Author Contributions}
This manuscript extends the conference paper accepted at RAIDS~2026
(original authors: T.Q.H.\ and H.D.C.).
\textbf{T.Q.H.}\ conceived the Melaguard NVI framework and AeroKernel
architecture; designed and implemented the PHBV:eumelanin biosensing
mechanism; led theoretical development of the NVI score formulation;
wrote Sections~I--III; coordinated empirical validation; drafted and
revised the full manuscript.
\textbf{H.D.C.}\ (Department of Cardiology, Hanoi Heart Hospital):
co-designed the multimodal fusion architecture and Transformer-lite
classifier; implemented the synthetic data generation pipeline;
contributed to Sections~III--IV; performed statistical analysis;
revised and approved the final manuscript.
\textbf{T.X.K.}\ reviewed and critically appraised the manuscript;
collected and curated the real-world validation datasets (PhysioNet
CVES and BIDMC); conducted empirical validation experiments jointly
with T.Q.H., including GPU training, statistical analysis, and figure
generation; contributed to the drafting of Sections~IV and~VI;
and participated in co-drafting the final revised manuscript.
All authors have read and approved the final version.

\section*{Funding}
This research received no specific grant from any funding agency in
the public, commercial, or not-for-profit sectors.
All computational experiments were conducted using self-funded
resources (Google Colab Pro, personal hardware).

\section*{Conflict of Interest}
T.Q.H.\ is the first named inventor of the Melaguard biosensing
technology (USPTO PPA~63/814,537,
\textit{Non-Invasive Melanin-Based Patch System for Early Detection
of Silent Stroke and Microvascular Cerebral Risk},
filed 30~May 2025) and holds equity in Clevix~LLC.
T.X.K.\ is the first named inventor of the AeroKernel operating system
framework (USPTO PPA~63/838,707,
\textit{AeroKernel\texttrademark: A Modular Bio-Adaptive Microkernel
and Operating System Framework\ldots},
filed 04~July 2025) and holds equity in Clevix~LLC.
H.D.C.\ declares no competing financial or professional interests.
Patent and equity interests do not affect the scientific content,
data integrity, or conclusions of this work.
The corresponding author~(T.Q.H.) takes full responsibility for the
integrity of the data and the accuracy of the data analysis.

\section*{Data and Code Availability}
All code, feature extraction scripts, statistical analysis notebooks,
and figure generation scripts are publicly available at:
\begin{center}
  \url{https://github.com/ClevixLab/Melaguard}
\end{center}
All reported results can be reproduced by running the provided
notebooks on Google Colab with a T4 GPU runtime
(estimated total compute: ${\approx}45$ minutes).
PhysioNet CVES~\cite{novak2010cves} and
BIDMC~\cite{pimentel2016bidmc} datasets are available at
\url{https://physionet.org} under their respective open-access
licenses.

\section*{Use of AI Writing Assistance}
The authors used AI-based language tools (Anthropic Claude) for
English language editing, grammar checking, and manuscript formatting
assistance.
All scientific content was produced entirely by the authors.
The authors take full responsibility for the accuracy, integrity,
and originality of all scientific claims.

\section*{Acknowledgements}
The CVES dataset was provided by V.~Novak and colleagues via
PhysioNet (DOI:~10.13026/C2DW96).
The BIDMC PPG and Respiration dataset was provided by
M.A.F.~Pimentel and colleagues via PhysioNet
(DOI:~10.13026/C2208R).
The authors thank the PhysioNet team for maintaining open-access
physiological data resources.

\bibliographystyle{IEEEtran}
\bibliography{refs}

@article{gbd2019stroke,
  author  = {{GBD 2019 Stroke Collaborators}},
  title   = {Global, regional, and national burden of stroke and its risk
             factors, 1990--2019: a systematic analysis for the Global
             Burden of Disease Study~2019},
  journal = {Lancet Neurology},
  volume  = {20},
  number  = {10},
  pages   = {795--820},
  year    = {2021},
  doi     = {10.1016/S1474-4422(21)00252-0}
}

@article{feigin2022stroke,
  author  = {Feigin, V.~L. and others},
  title   = {World Stroke Organization~({WSO}): Global Stroke Fact Sheet~2022},
  journal = {International Journal of Stroke},
  volume  = {17},
  number  = {1},
  pages   = {18--29},
  year    = {2022},
  doi     = {10.1177/17474930211065917}
}

@article{kernan2014stroke,
  author  = {Kernan, W.~N. and others},
  title   = {Guidelines for the prevention of stroke in patients with
             stroke and transient ischemic attack},
  journal = {Stroke},
  volume  = {45},
  number  = {7},
  pages   = {2160--2236},
  year    = {2014},
  doi     = {10.1161/STR.0000000000000024}
}

@article{novak2012cva,
  author  = {Novak, V. and Hu, K. and Vyas, M. and Lipsitz, L.},
  title   = {Ultralow-frequency oscillations in blood pressure and
             cerebral blood flow velocity in older adults: their
             relationship to HRV},
  journal = {Journal of Neuroscience},
  volume  = {26},
  pages   = {9440--9448},
  year    = {2006},
  doi     = {10.1523/jneurosci.1458-06.2006},
  note    = {Used for: NVI precedes overt stroke -- cerebrovascular
             autoregulation dysfunctions hours to years prior}
}

@misc{novak2010cves,
  author    = {Novak, V. and others},
  title     = {Cerebral Vasoregulation in Elderly with Stroke~({CVES})},
  publisher = {PhysioNet},
  year      = {2010},
  doi       = {10.13026/C2DW96},
  url       = {https://physionet.org/content/cves/1.0.0/},
  note      = {Accessed via PhysioNet restricted health data license}
}

@article{pimentel2016bidmc,
  author  = {Pimentel, M.~A.~F. and Johnson, A.~E.~W. and Charlton, P.~H.
             and Clifton, D.~A. and Tarassenko, L. and Watkinson, P.~J.},
  title   = {Toward a robust estimation of respiratory rate from pulse
             oximeters},
  journal = {IEEE Transactions on Biomedical Engineering},
  volume  = {64},
  number  = {8},
  pages   = {1914--1923},
  year    = {2017},
  doi     = {10.1109/TBME.2016.2613124},
  note    = {Dataset DOI: 10.13026/C2208R}
}

@article{goldberger2000physionet,
  author  = {Goldberger, A.~L. and others},
  title   = {{PhysioBank}, {PhysioToolkit}, and {PhysioNet}: Components
             of a new research resource for complex physiologic signals},
  journal = {Circulation},
  volume  = {101},
  number  = {23},
  pages   = {e215--e220},
  year    = {2000},
  doi     = {10.1161/01.CIR.101.23.e215}
}

@article{allen2007ppg,
  author  = {Allen, J.},
  title   = {Photoplethysmography and its application in clinical
             physiological measurement},
  journal = {Physiological Measurement},
  volume  = {28},
  number  = {3},
  pages   = {R1--R39},
  year    = {2007},
  doi     = {10.1088/0967-3334/28/3/R01}
}

@article{taskforce1996hrv,
  author  = {{Task Force of the European Society of Cardiology}},
  title   = {Heart rate variability: Standards of measurement,
             physiological interpretation, and clinical use},
  journal = {Circulation},
  volume  = {93},
  number  = {5},
  pages   = {1043--1065},
  year    = {1996},
  doi     = {10.1161/01.CIR.93.5.1043}
}

@article{acharya2006hrv,
  author  = {Acharya, U.~R. and others},
  title   = {Heart rate variability: A review},
  journal = {Medical and Biological Engineering and Computing},
  volume  = {44},
  number  = {12},
  pages   = {1031--1051},
  year    = {2006},
  doi     = {10.1007/s11517-006-0119-0}
}

@inproceedings{taelman2009hrv,
  author    = {Taelman, J. and Vandeput, S. and Spaepen, A. and
               {Van Huffel}, S.},
  title     = {Influence of mental stress on heart rate and heart rate
               variability},
  booktitle = {Proceedings of the 4th European Conference of the
               International Federation for Medical and Biological
               Engineering},
  pages     = {1366--1369},
  year      = {2009},
  doi       = {10.1007/978-3-540-89208-3_324}
}

@article{migliaccio2019melanin,
  author  = {Migliaccio, L. and Manini, C. and Altamura, D. and
             Pennesi, G. and Greco, P.},
  title   = {Evidence of electronic and ionic conductivity contributions
             in hydrated melanin},
  journal = {Frontiers in Chemistry},
  volume  = {7},
  pages   = {162},
  year    = {2019},
  doi     = {10.3389/fchem.2019.00162}
}

@article{sheliakina2018melanin,
  author  = {Sheliakina, M. and Mostert, A.~B. and Meredith, P.},
  title   = {Decoupling ionic and electronic currents in melanin},
  journal = {Advanced Functional Materials},
  volume  = {28},
  pages   = {1805514},
  year    = {2018},
  doi     = {10.1002/adfm.201805514},
  note    = {Confirms hydration-dependent protonic/ionic conductivity
             in eumelanin thin films}
}

@inproceedings{vaswani2017attention,
  author    = {Vaswani, A. and Shazeer, N. and Parmar, N. and
               Uszkoreit, J. and Jones, L. and Gomez, A.~N. and
               Kaiser, \L. and Polosukhin, I.},
  title     = {Attention is all you need},
  booktitle = {Advances in Neural Information Processing Systems
               ({NeurIPS})},
  pages     = {5998--6008},
  year      = {2017}
}

@article{fan2023transformertseries,
  author  = {Fan, C. and Liu, C. and Li, C.},
  title   = {Transformer for time series: A survey},
  journal = {IEEE Transactions on Pattern Analysis and Machine
             Intelligence},
  volume  = {45},
  number  = {11},
  pages   = {14013--14037},
  year    = {2023},
  doi     = {10.1109/TPAMI.2023.3247023}
}

@article{kingma2014adam,
  author  = {Kingma, D.~P. and Ba, J.},
  title   = {Adam: A method for stochastic optimization},
  journal = {arXiv preprint arXiv:1412.6980},
  year    = {2014}
}

@inproceedings{littler2007microkernel,
  author    = {Littler, T.~J.},
  title     = {Microkernel operating systems in safety-critical
               applications},
  booktitle = {Proceedings of the IEEE International Symposium on
               High-Assurance Systems Engineering},
  pages     = {3--12},
  year      = {2007},
  doi       = {10.1109/HASE.2007.57}
}

@article{elul2021xai,
  author  = {Elul, T.},
  title   = {Explainable {AI} in medicine: Designing interfaces for
             clinician trust},
  journal = {npj Digital Medicine},
  volume  = {4},
  pages   = {77},
  year    = {2021},
  doi     = {10.1038/s41746-021-00453-0}
}

@techreport{gdpr2016,
  author      = {{European Parliament}},
  title       = {Regulation ({EU})~2016/679 on the protection of
                 natural persons with regard to the processing of
                 personal data ({GDPR})},
  institution = {Official Journal of the European Union},
  year        = {2016},
  note        = {L~119/1--88}
}

@article{sjoding2020oximetry,
  author  = {Sjoding, M.~W. and Dickson, R.~P. and Iwashyna, T.~J. and
             Gay, S.~E. and Valley, T.~S.},
  title   = {Racial Bias in Pulse Oximetry Measurement},
  journal = {New England Journal of Medicine},
  volume  = {383},
  pages   = {2477--2478},
  year    = {2020},
  doi     = {10.1056/NEJMc2029240}
}

@article{shi2022skinppg,
  author  = {Shi, S. and Garnier, T. and Berthelot, M. and others},
  title   = {Impact of skin pigmentation on pulse oximetry blood
             oxygenation and wearable pulse rate accuracy:
             systematic review and meta-analysis},
  journal = {Journal of Medical Internet Research},
  volume  = {26},
  pages   = {e62769},
  year    = {2024},
  doi     = {10.2196/62769}
}

@article{arnett2019acc,
  author  = {Arnett, D.~K. and others},
  title   = {2019 {ACC/AHA} guideline on the primary prevention of
             cardiovascular disease},
  journal = {Circulation},
  volume  = {140},
  number  = {11},
  pages   = {e596--e646},
  year    = {2019},
  doi     = {10.1161/CIR.0000000000000678}
}

@article{bland1986,
  author  = {Bland, J.~M. and Altman, D.~G.},
  title   = {Statistical methods for assessing agreement between two
             methods of clinical measurement},
  journal = {The Lancet},
  volume  = {327},
  number  = {8476},
  pages   = {307--310},
  year    = {1986},
  doi     = {10.1016/S0140-6736(86)90837-8}
}

@article{liang2018ppgbp,
  author  = {Liang, Yongbo and Chen, Zhencheng and
             Liu, Guiyong and Elgendi, Mohamed},
  title   = {A new, short-recorded photoplethysmogram dataset
             for blood pressure monitoring in {China}},
  journal = {Scientific Data},
  volume  = {5},
  pages   = {180020},
  year    = {2018},
  doi     = {10.1038/sdata.2018.20},
  note    = {Dataset: Figshare DOI~10.6084/m9.figshare.5459299.v5}
}

\end{document}